\documentclass[sigconf]{acmart}
\usepackage{makecell}

\usepackage{bm}
\usepackage{multirow}
\usepackage{booktabs}
\usepackage{threeparttable}
\usepackage{hyperref}
\usepackage{enumerate}
\usepackage{multirow}
\usepackage{subfig}

\AtBeginDocument{%
  \providecommand\BibTeX{{%
    \normalfont B\kern-0.5em{\scshape i\kern-0.25em b}\kern-0.8em\TeX}}}

\setcopyright{acmcopyright}
\copyrightyear{2021}
\acmYear{2021}

\acmConference{ACM ICAIF 2021 Workshop on Explainable AI in Finance}{November 3, 2021}{Virtual Conference}

\begin{document}

\title{Designing Inherently Interpretable Machine Learning Models}

\author{Agus Sudjianto and Aijun Zhang}
\authornote{The views expressed in this paper are the personal views of the authors and do not necessarily reflect the views of Wells Fargo Bank, N.A., its parent company, affiliates, and subsidiaries.}
\email{{agus.sudjianto, aijun.zhang}@wellsfargo.com}
\affiliation{%
  \institution{Corporate Model Risk, Wells Fargo, USA}
\country{}
}


\renewcommand{\shortauthors}{Sudjianto and Zhang}

\begin{abstract}
Interpretable machine learning (IML) becomes increasingly important in highly regulated industry sectors related to the health and safety or fundamental rights of human beings. 
In general, the inherently IML models should be adopted because of their transparency and explainability, while black-box models with model-agnostic explainability can be more difficult to defend under regulatory scrutiny. 
For assessing inherent interpretability of a machine learning model, we propose a qualitative template based on feature effects and model architecture constraints. It provides the design principles for high-performance IML model development, with examples given by reviewing our recent works on ExNN, GAMI-Net, SIMTree, and the Aletheia toolkit for local linear interpretability of deep ReLU networks. 
We further demonstrate how to design an interpretable ReLU DNN model with evaluation of conceptual soundness for a real case study of predicting credit default in home lending.  
We hope that this work will provide a practical guide of developing inherently IML models in high risk applications in banking industry, as well as other sectors. 
\end{abstract}



\maketitle

\section{Introduction}
Technological advances in artificial intelligence (AI) bring us both potential benefits and risks in solving real-world problems. While the machine learning algorithms become more complex and built into end-to-end automation, they become more difficult to understand by human beings. There is increasing public concern about the risk of misusing AI and machine learning algorithms, especially in highly-risk applications related to the health and safety or fundamental rights of human beings. A recent Harvard Business Review article titled ``AI Regulation Is Coming'' \cite{HBR2021} explains why the governments move to regulate AI systems with legal requirements on fairness, transparency and explainability. For more information about AI regulation, one may refer to the Wikipedia item \cite{Wiki2021} and European Union's 2021 proposal for an AI legal framework \cite{EU2021}.

For banking and finance as a highly regulated industry, the US agencies recently issued a request for information and comment on financial institutions' use of AI (including machine learning) \cite{RFI2021}. Among the requested questions, the top three  are related to lack of explainability in AI approaches, and there are another five questions related to fair lending (which often addresses model transparency). A less transparent and explainable approach may cause difficulty for conceptual soundness evaluation, a critical expectation for model risk management as prescribed in SR11-7/OCC11-12 \cite{SR11-7}.
In August 2021, the OCC published the handbook on model risk management \cite{OCC2021}, which states explicitly that ``if the bank uses AI models, examiners should assess if model ratings take explainability into account.'' It adds that ``transparency and explainability are key considerations that are typically evaluated as part of effective risk management regarding the use of complex models.''

Black-box AI models are by definition neither transparent nor explainable. Model-agnostic methods like LIME (locally interpretable model-agnostic explanations) \cite{LIME2016} and SHAP (SHapley Additive exPlanations) \cite{SHAP2017} provide approximate explanations, but have many potential pitfalls as elaborated by \cite{molnar2020pitfalls}. Analogous to the no free lunch theorem for optimization and machine learning, there does not exist such a model-agnostic one-fits-all explainability.  One may refer to Figure~2 of \cite{molnar2020pitfalls} for an interesting example about how erroneous the XGBoost (eXtreme Gradient Boosting) and SHAP results could be. There are indeed significant amount of academic papers criticizing the blind use of model-agnostic explanation methods \cite{rudin2019stop, slack2020fooling, kumar2020problems} (to name a few).

In this paper we focus on inherently interpretable machine learning (IML) models with model-based explainability.  Throughout the paper, IML means inherently interpretable machine learning, where inherent interpretability means that model is transparent and self-explanatory. It is argued by \cite{yang2020enhancing} that the inherent interpretability of a complex model ought to be induced from practical constraints. We extend this paradigm for qualitatively assessing the interpretability of IML models, which in turn provides design principles for IML model development from feature-wise and model-wise perspectives. The design principles are exemplified with our recently developed high-performance IML models:
\begin{itemize}
\item {\bf ExNN} (enhanced explainable neural networks) \cite{yang2020enhancing}; 
\item {\bf GAMI-Net} (generalized additive models with structured interactions) \cite{yang2021gami} ;
\item {\bf SIMTree} (single-index model tree) \cite{SIMTree2021};
\item {\bf Aletheia} toolkit for local linear interpretability of deep ReLU networks  \cite{sudjianto2020unwrapping}. 
\end{itemize}
To demonstrate how to design an IML model based on deep ReLU networks, we also
present a real world case study of predicting credit default for home lending, where the Aletheia toolkit is employed for model interpretation, diagnostics, and simplification. We declare that this review paper is largely based on four of our published papers cited above. It is our hope that this work will provide a practical guide of developing inherently IML models in high risk applications in banking industry, as well as other public or private sectors. 

 
\section{Model Interpretability}
Model interpretability is a loosely defined concept and does not have a common {\em quantitative} measurement \cite{murdoch2019definitions}. Here we follow the interpretability constraint paradigm from \cite{yang2020enhancing} and attempt to propose a {\em qualitative} template for model-based interpretability assessment. For simplicity of discussion, we consider only the tabular data modeling in what follows.

Consider a machine learning model with multi-dimensional inputs. Each input is called a feature, and its functional relationship with the final prediction is called a feature effect. When  each feature relates with the final prediction independently, it is easy to interpret the univariate feature effect (e.g. a line or a curve) per feature, and each feature's importance can be measured or rank-ordered.  Such additive decomposition of feature effects corresponds to the generalized linear model (GLM) and generalized additive model (GAM).
For better interpretability, the model is desired to contain as small number of features as possible, and the feature effects are desired to be as smooth as possible. The characteristics that a model uses only a small subset of features is called sparsity. 

In most cases, however, the features have joint effects to the final prediction, e.g. two-way or multi-way interactions. A two-way interaction effect can be directly modeled as a bivariate function, which can be visualized by a surface plot or heatmap for visual interpretation by human beings. A multi-way interaction modeled by a generic multivariate function is difficult to interpret as we live in a 3D world. In statistics and machine learning, projection is such a modeling strategy to combine the multiple inputs into a single-dimensional artificial feature, then apply the ridge function to capture the joint effect. For simplicity, linear projection is often considered. This corresponds to the single-index model (SIM), and such linear projection also appears in the projection layer of deep neural networks (DNN). 
For better interpretability, sparse projection with a small subset of features is preferred; when multiple projections take place, e.g. as in GAIM (generalized additive index model) and DNN, the near-orthogonality of projection weights are sometimes desired, as that would clear up the confounding in effect interpretation.

In other cases when the data is heterogeneous, the same feature may have different effects for different data clusters. Segmentation is an effective strategy commonly used for practical data modeling. An ad hoc way is manually dividing data into different segments, then fit each segment of data with a separate model. The more intelligent approaches include decision trees (recursively partitioning data into segments), model-based trees, and neural networks with ReLU activation. 
For better interpretability, smaller number of segments are desired. 

Besides, model interpretation ought to comport with expert knowledge in the application domain. Domain experts are built upon prior experience and wisdom, and they can provide valuable advice for IML model building. For example, certain features are advised to be modeled monotonically increasing or decreasing, then an IML model is desired to be capable to take such monotonicity constraints into account.  


Thus we have described multiple interpretability constraints, including additivity, sparsity, linearity, smoothness, monotonicity, and near-orthogonality. These constraints tend to make the model interpretable by human beings. To understand model-based feature effects and feature importance are our primary interpretation purposes. There are two types of interpretability often used in the IML context: 
\begin{itemize}
\item Local interpretability: feature effects and importance are evaluated based on a single individual point or its small neighborhood; 
\item Gobal interpretability: feature effects and importance are evaluated on the entire dataset, for the purpose of understanding the overall model behavior. 
\end{itemize} 
It is reasonable to claim that the segmented modeling would trade-off between local and global interpretability. In general, it is desirable to have an IML model with smaller number of segments, while each segment can also be locally explainable. 

To sum up, Table~\ref{tab:template} provides a qualitative template for assessing the model-based interpretability. It also comes with two rating columns for local and global assessments of each interpretability constraints, with a four-star rating mechanism:
\begin{itemize}
\item[] $\ast\ast\ast\ast$ - model automatically satisfies the constraint;
\item[] $\ast\ast\ast$ - model can be easily constrained;
\item[] $\ast\ast$ - model can be roughly constrained;
\item[] $\ast$ - model cannot satisfy the constraint.
\end{itemize}
The ratings in Table~\ref{tab:template} are given for the GLM as a show case, where GLM does not perform projection and segmentation. Note that ``NA'' for projection or segmentation means that the strategy is not used, so there is additional complexity so caused for the interpretation purpose.   
The bottom row is the overall assessment of local interpretability and global interpretability, also rated with one to four stars, with four-star indicating the (nearly) perfect interpretability, and one-star indicating really poor interpretability. 

\begin{table*}[t]
\caption{Qualitative Assessment Template for Model Interpretability (with ratings given for GLM)}\label{tab:template}
\centering
\setlength{\tabcolsep}{2pt}
\renewcommand{\arraystretch}{1.2}
\begin{tabular}{| c | c | c | c | }\hline 
Model Characteristics & Description & Local Rating & Global Rating \\\hline\hline
Additivity &  \makecell{Whether/how model takes the additive or modular form\\
	(Additive decomposition of feature effects tends to be more interpretable) } 
	& **** & **** \\\hline  
Sparsity & \makecell{Whether/how features or model components are regularized\\
	(Having fewer features or components tends to be more interpretable) } 
	& *** & *** \\\hline  
Linearity & \makecell{Whether/how feature effects are linear (constant as special case)\\
	(Linear or constant feature effects are easy to interpret) } 
	& **** & **** \\\hline  
Smoothness & \makecell{Whether/how feature effects are continuous and smooth\\
	(Continuous and smooth feature effects are relatively easy to interpret) } 
	& **** & **** \\\hline  
Monotonicity & \makecell{Whether/how feature effects can be modeled to be monotone\\
	(When increasing/decreasing effects are desired by expert knowledge)} 
	& *** & *** \\\hline  
Visualizability & \makecell{Whether/how the feature effects can be directly visualized\\
	(Visualization facilitates the final model diagnostics and interpretation)} 
	& **** & **** \\\hline\hline
Projection &   \makecell{Whether/how projection is used for feature combination\\
	(Sparse and near-orthogonal projection tends to be more interpretable) } &  
	NA & NA \\\hline  
Segmentation &   \makecell{Whether/how segmentation is used for heterogeneous modeling\\ 
	(Having smaller number of segments tends to be more interpretable)}  & 
	NA &  NA \\\hline\hline
Others & Placeholder for other desired interpretability characteristics &  &   \\\hline\hline
Overall Interpretability & Summary statement  &  4-star  & 4-star  \\\hline
\end{tabular}
\end{table*}
   
With such a qualitative assessment template, we may rate GAM with 3.5-star both local and global interpretability, and rate decision tree with 4-star local interpretability and 2.5-star global interpretability. In general, these classical statistical models have relatively high interpretability, while their predictive performance is relatively limited when modeling complex data. Next we discuss how to design high-performance IML models with extended model architectures subject to interpretability constraints.

\section{Designing High-Performance IML Models}
High-performance ML models for tabular data include tree ensembles (e.g. random forest and gradient boosting machine) and deep neural networks (e.g. DNN and ResNet). However, these models are known of lacking of transparency and explainability, except for DNN with ReLU activation that was studied by \cite{sudjianto2020unwrapping} to achieve a certain level of transparency.  In this section, we discuss how to design high-performance ML models with inherent interpretability.

We suggest to design an IML model from a) feature selection and b) model architecture perspectives. Model interpretability relies on the feature effects and importance, while the features themselves need to be interpretable as well. Interpretable feature selection may come from the following considerations: 
\begin{itemize}
\item Meaningful feature with real meaning in practice;
\item Monotonic feature according to expert knowledge;
\item Causal feature from causal discovery;
\item Interpretable feature engineering (e.g. sparse projection).
\end{itemize}

To ensure high-performance, we suggest to start with model architectures with rich representation power, e.g. DNN or GAIM with universal approximation capability. Then, we may inject model-based interpretability by imposing the architecture constraints; cf. Table~\ref{tab:template}. This is a design principle for high-performance IML models we have developed in the past couple of years, which are briefly reviewed below. 

\subsection{Enhanced Explainable Neural Networks}
\begin{figure}[!htb]
  \centering
  \includegraphics[width=\linewidth]{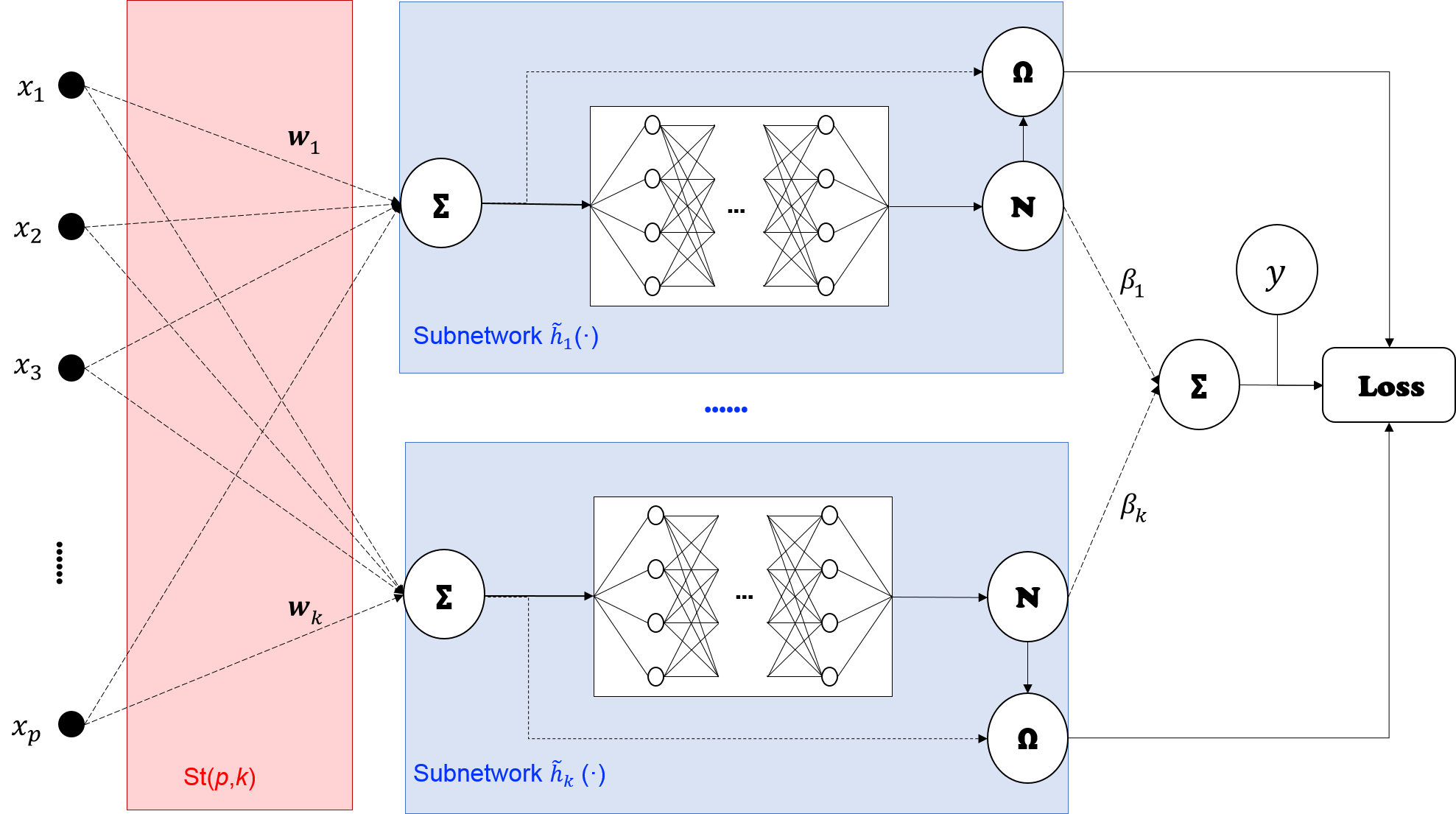}
  \caption{The ExNN architecture \cite{yang2020enhancing}, subject to a) sparsity constraints  on both projection layer and output layer, b) orthogonality constraint on projection layer layer, and c) smoothness constraint on each ridge function as modeled by a subnetwork. } \label{fig:ExNN}
\end{figure}

The enhanced explainable neural networks (ExNN) is designed to integrate GAIM and DNN. The model interpretability are induced from the sparsity, orthogonality and smoothness constraints.  Figure~\ref{fig:ExNN} shows the model architecture with such interpretability constraints. It was shown with extensive numerical examples in \cite{yang2020enhancing} that ExNN has a superior balance between prediction performance and model interpretability. More details about ExNN can be referred to our paper \cite{yang2020enhancing} recently published  in {\em IEEE Trans. on Neural Networks and Learning Systems}.

Our assessment of model-based interpretability for ExNN is provided in Table~\ref{tab:TempExNN}. 

\begin{table}[!htb]
\caption{Interpretability Assessment of ExNN  \cite{yang2020enhancing} }\label{tab:TempExNN}
\centering
\setlength{\tabcolsep}{2pt}
\renewcommand{\arraystretch}{1.2}
\begin{tabular}{| c | c | c |}\hline 
Characteristics &  Local Rating & Global Rating \\\hline\hline
Additivity 	& **** & **** \\\hline  
Sparsity 	& *** & *** \\\hline  
Linearity  & * & * \\\hline  
Smoothness  & *** & *** \\\hline  
Monotonicity & **  & ** \\\hline  
Visualizability & **** & **** \\\hline\hline  
Projection  & *** & *** \\\hline  
Segmentation  & NA &  NA \\\hline\hline
Overall Interpretability  &  3-star  & 3-star  \\\hline
\end{tabular}
\end{table}

\subsection{GAMI-Net}  
The GAMI-Net is an explainable neural network based on generalized additive models with structured interactions. It is a disentangled feedforward network with additive subnetworks capturing each of main effects and pairwise interactions. Three interpretability constraints are imposed for GAMI-Net, including sparsity, heredity and marginal clarity.  It was shown in \cite{yang2021gami} that GAMI-Net is pretty explainable while achieving competitive prediction accuracy in comparison to the explainable boosting machine and other classic machine learning models. 
More details about GAMI-Net can be referred to our paper  \cite{yang2021gami}  recently published  in {\em Pattern Recognition}.

\begin{figure}[!htb]
  \centering
  \includegraphics[width=\linewidth]{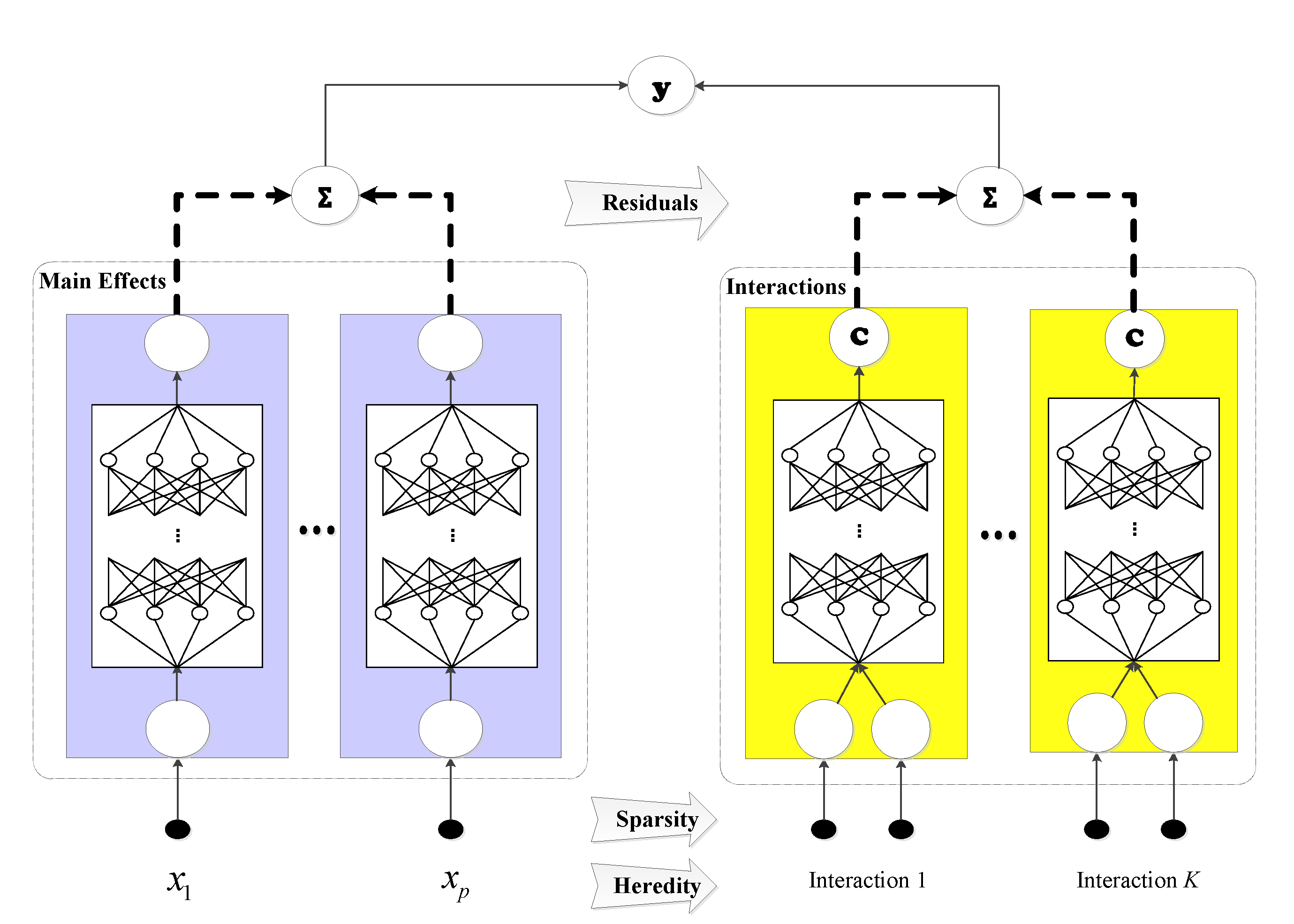}
  \caption{
  The GAMI-Net architecture \cite{yang2021gami}, subject to a) sparsity constraints on the number of main effects and pairwise interactions, b) heredity on the candidate pairwise interactions; and c) marginal clarity to avoid confusion between main effects and their child pairwise interactions. } \label{fig:GAMI-Net}
\end{figure}

Our assessment of model-based interpretability for GAMI-Net is provided in Table~\ref{tab:TempGAMI-Net}.  Note that the local rating for linearity ($\ast$ or $\ast\ast\ast\ast$)  depends on whether the ReLU activation function is used for the subnetworks. In \cite{sudjianto2020unwrapping}, we have studied that deep ReLU networks can be unwrapped to a set of local linear models, so its local feature effect is linear.  

\begin{table}[!htb]
\caption{Interpretability Assessment of GAMI-Net \cite{yang2021gami}}
\label{tab:TempGAMI-Net}
\centering
\setlength{\tabcolsep}{2pt}
\renewcommand{\arraystretch}{1.2}
\begin{tabular}{| c | c | c |}\hline 
Characteristics &  Local Rating & Global Rating \\\hline\hline
Additivity 	& **** & **** \\\hline  
Sparsity 	& *** & *** \\\hline  
Linearity  & */**** & * \\\hline  
Smoothness  & *** & *** \\\hline  
Monotonicity & **  & ** \\\hline  
Visualizability & **** & **** \\\hline\hline
Projection  & NA & NA \\\hline  
Segmentation  & NA &  NA \\\hline\hline
Overall Interpretability  &  3.5-star  & 3.5-star  \\\hline
\end{tabular}
\end{table}

\subsection{SIMTree}
\begin{figure}[!htb]
  \centering
  \includegraphics[width=\linewidth]{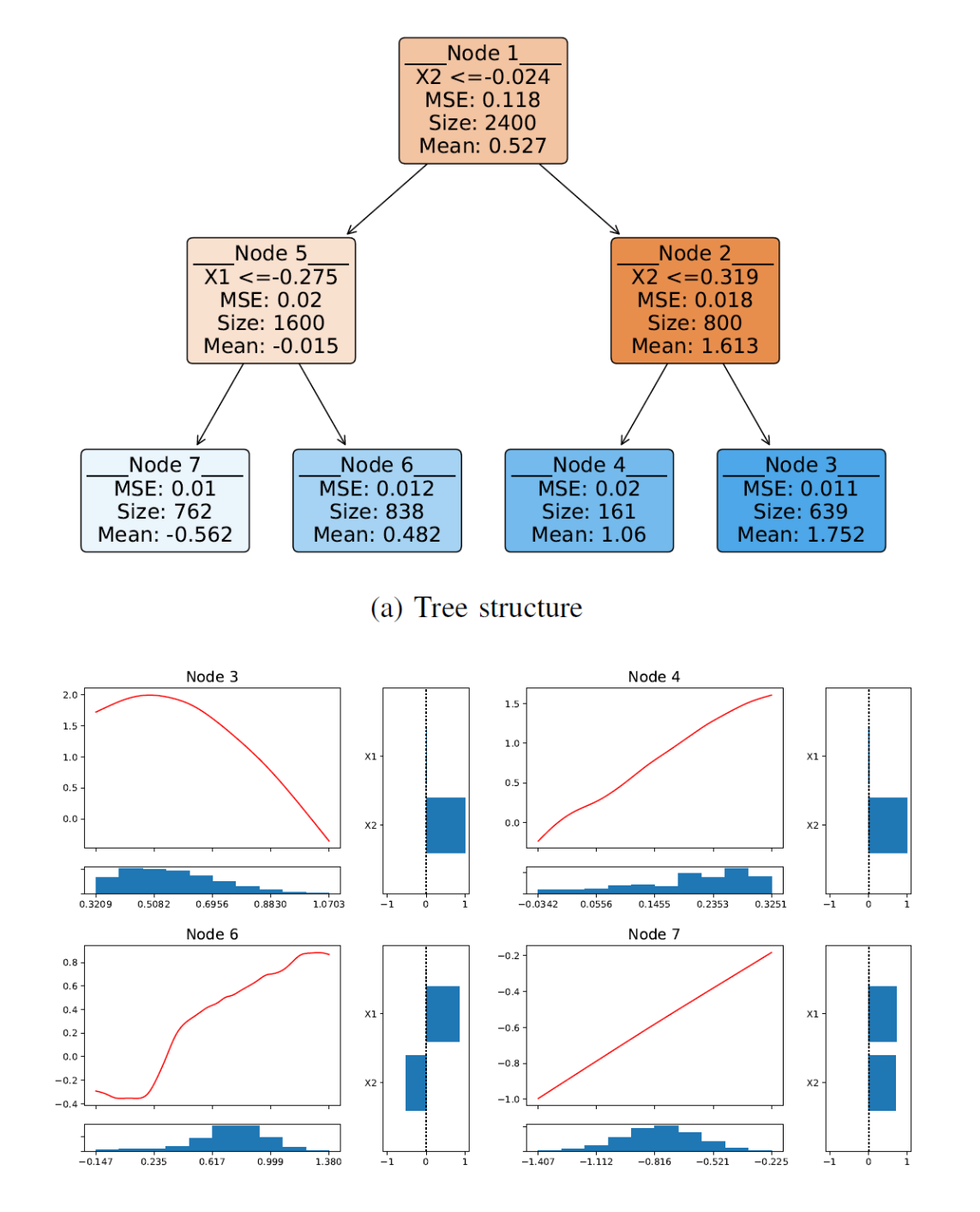}
  \caption{
  The SIMTree architecture that uses single-index models for each terminal node \cite{SIMTree2021}.   } \label{fig:SIMTree}
\end{figure}

The SIMTree (single-index model tree) is a model-based tree that uses inherently interpretable SIM model for each terminal node. It is a natural extension of the standard decision tree with recursive partitioning. For each segment data at the terminal node, the choice of SIM would increase prediction performance while maintaining a decent level of interpretability.  See Figure~\ref{fig:SIMTree} for an illustration of SIMTree model structure. 
More details about SIMTree can be referred to our paper \cite{SIMTree2021}  to be published  in {\em IEEE Trans. on Knowledge and Data Engineering}.

Our assessment of model-based interpretability for SIMTree is provided in Table~\ref{tab:TempSIMTree}. 

\begin{table}[!htb]
\caption{Interpretability Assessment of SIMTree \cite{SIMTree2021}}
\label{tab:TempSIMTree}
\centering
\setlength{\tabcolsep}{2pt}
\renewcommand{\arraystretch}{1.2}
\begin{tabular}{| c | c | c |}\hline 
Characteristics &  Local Rating & Global Rating \\\hline\hline
Additivity 	& **** & ** \\\hline  
Sparsity 	& *** & ** \\\hline  
Linearity  & * & * \\\hline  
Smoothness  & *** & * \\\hline  
Monotonicity & ***  & * \\\hline  
Visualizability & **** & *** \\\hline\hline
Projection  & *** & NA \\\hline  
Segmentation  & NA &  *** \\\hline\hline
Overall Interpretability  &  4-star  & 2.5-star  \\\hline
\end{tabular}
\end{table}

\subsection{Sparse ReLU DNNs}
The deep neural networks (DNNs) with ReLU activation functions can be unwrapped into a set of local linear models (LLMs) for disjoint data segments, and they can be interpreted with model diagnostics and simplification by using the Aletheia toolkit developed by \cite{sudjianto2020unwrapping}. Among other ways of model simplification, we may impose sparsity constraint on the neuron weights, so that the ReLU DNN would obliquely partition the sample space into smaller number of segments, thus increasing the level of model interpretability. 

\begin{figure}[!htb]
  \centering
  \includegraphics[width=\linewidth, height=0.6\linewidth]{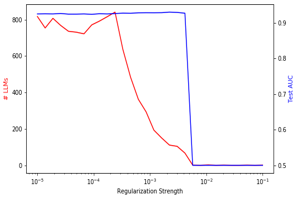}
  
  \includegraphics[width=\linewidth]{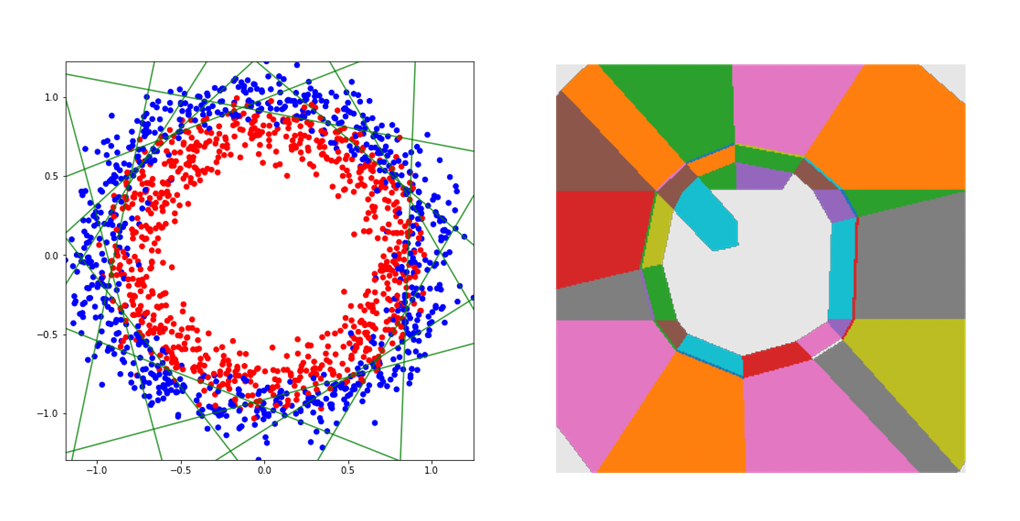}
  \caption{Sparse ReLU DNN with local exact interpretability, illustrated through a two-dimensional CoCircles data \cite{sudjianto2020unwrapping}. (a) Prediction performance and the number of unwrapped  LLMs subject to varying degrees of $\ell_1$-sparsity constraint. (b) The resulting non-trivial LLMs and space partitioning upon an appropriate choice of $\ell_1$ hyperparameter.} \label{fig:SparseDNN}
\end{figure}

Figure~\ref{fig:SparseDNN} shows the sparse ReLU DNN with local exact interpretability for a two-dimensional CoCircles data as in \cite{sudjianto2020unwrapping}. It can be found that the unwrapped LLMs form a reasonable approximation of the underlying decision boundary (i.e. a circle). By sparsity constraint, the deep ReLU networks are simplified to 10+ non-trivial LLMs, which is relatively easy to interpret. 

Our assessment of model-based interpretability for Sparse ReLU DNNs is provided in Table~\ref{tab:TempSparseDNN}. 

\begin{table}[!htb]
\caption{Interpretability Assessment of Sparse ReLU DNNs.}
\label{tab:TempSparseDNN}
\centering
\setlength{\tabcolsep}{2pt}
\renewcommand{\arraystretch}{1.2}
\begin{tabular}{| c | c | c |}\hline 
Characteristics &  Local Rating & Global Rating \\\hline\hline
Additivity 	& **** & ** \\\hline  
Sparsity 	& *** & *** \\\hline  
Linearity  & **** & * \\\hline  
Smoothness  & **** & *** \\\hline  
Monotonicity & ***  & ** \\\hline  
Visualizability & **** & ** \\\hline\hline
Projection  & NA & *** \\\hline  
Segmentation  & NA &  *** \\\hline\hline
Overall Interpretability  &  4-star  & 2.5-star  \\\hline
\end{tabular}
\end{table}

%

\section{Case Study on Home Lending Dataset} 
In this section, we present a real world case study of predicting credit default for home lending.  In a regulated financial institution, model interpretability is a requirement for the purpose of evaluating the model's conceptual soundness  as well as explaining to model users or customers the decision made by model. Part of conceptual soundness requirement is that the feature effects must be consistent with the expected business knowledge. For example, lower credit quality (e.g. FICO score) drives higher probability of default. 

The lack of ability to evaluate the conceptual soundness of DNNs has been a major obstacle for credit risk applications.  The purpose of this case study is to illustrate how we may design an IML model based on deep ReLU networks, using the Aletheia toolkit developed by \cite{sudjianto2020unwrapping}. By LLM-based diagnostics and simplification, we can provide strong mechanism to evaluate model conceptual soundness of ReLU DNNs equivalent to statistical linear regression models. 

We take a sample of 16,000 residential mortgage loans between the period of 2004 to 2014. In this panel data, each account is observed over multiple times so correlation exists in the response variable. We train a ReLU DNN on data split according to loan IDs so that a specific loan ID is exclusive either in training, validation or testing sets to avoid overfitting due to correlation among training and validation/testing data sets. The response variable of interest is binary with the label 0 as non-default and 1 as default, while the features include various loan characteristics and other auxiliary variables.  A partial list of 19 features are summarized in Table~\ref{Homelending_Introduction}.  Note that the original dataset is imbalanced in terms of binary responses, where only a small fraction (less than 1\%) are defaulting cases. For simplicity of illustration, we created a balanced sample data set with  8000 default cases and 8000 non-default cases. 

\begin{table}[!h]
\begin{center}
\caption{Partial list of features and their definition in the Homelending Dataset.} \label{Homelending_Introduction}
\begin{tabular}{ll}\hline\hline
{\bf Variable}    &   {\bf Meaning}  \\ \hline
balloon\_in & Indicator for Balloon Payment\\
delq0 & Delinquency status at snapshot time\\
fico0 &  Refreshed FICO at snapshot time\\
iARM & Indicator for ARM Loan   \\
iCashOut & Indicator for Cash Out\\
iFullDoc & Indicator for Full Doc\\
interest{\_}only{\_}in & Indicator for Interest Only \\ 
iOptARM & Indicator for Option Arm Product\\
iSF & Indicator for Single Family\\
loanAmt & Current Loan Amount\\
loanPurp & Loan Purpose Code\\
ltv{\_}fcast   &   Loan to value ratio forecasted  \\ 
orig\_HPI & Origination HPI \\
orig\_int\_rt & Origination Interest Rate\\
orig\_ltv &	 Origination LTV\\
SATO2   &    Spread at Time of Origination  \\ 
TotEmpyy	& Total Employment YoY\\
TotPersIncyy & Total Personal Income YoY\\
crtl	& Unpaid Balance/loan amount\\
QSpread2	 & Note rate - mortgage rate\\
\hline\hline
\end{tabular}
\end{center}
\end{table}
	
	\subsection{Model Details}
	For illustration purposes, we use a ReLU net with two hidden layers (each layer with 5 neurons). This ReLU net model has a performance on the test set of 75.77\% accuracy and ROC AUC of 84.29\%, that are comparable to the performance of best models reported earlier by \cite{chen2020adaptive}.  By using the Aletheia toolkit\footnote{https://github.com/SelfExplainML/Aletheia},  we obtain the summary of the largest regions such as the number of samples and the performance of local linear models from the ReLU net, as shown in Table~\ref{Homelending_llms}. There are a total of 62 regions of which 14 (23\%) are single sample instances or a single class.
	
	\begin{table}[!h]
	\setlength{\tabcolsep}{2pt}
\renewcommand{\arraystretch}{1.2}
		\begin{center}
			\caption{Summary of 15 Largest Regions Unwrapped from ReLU-Net.} \label{Homelending_llms}
			\begin{tabular}{cccccc}
				\hline\hline
				& Count & Response Mean & Response Std & Local AUC & Global AUC \\ \hline
				0  & 2303 &    0.461     &    0.498     &  0.791   &   0.834    \\
				1  & 615  &    0.668     &    0.471     &  0.766   &   0.824    \\
				2  & 527  &    0.557     &    0.497     &  0.718   &   0.832    \\
				3  & 478  &    0.495     &    0.499     &  0.699   &   0.831    \\
				4  & 320  &    0.043     &    0.204     &  0.641   &   0.829    \\
				5  & 258  &    0.732     &    0.442     &  0.674   &   0.806    \\
				6  & 218  &    0.825     &    0.379     &  0.628   &   0.768    \\
				7  & 206  &    0.038     &    0.193     &  0.761   &   0.784    \\
				8  & 187  &    0.658     &    0.474     &  0.688   &   0.804    \\
				9  & 155  &    0.935     &    0.245     &  0.579   &   0.823    \\
				10 & 138  &    0.217     &    0.412     &  0.631   &   0.820    \\
				11 & 131  &    0.145     &    0.352     &  0.679   &   0.791    \\
				12 & 121  &    0.793     &    0.404     &  0.775   &   0.796    \\
				13 & 116  &    0.129     &    0.335     &  0.606   &   0.815    \\
				14 & 105  &    0.647     &    0.477     &  0.714   &   0.812    \\
				15 & 103  &    0.106     &    0.308     &  0.590   &   0.761    \\ \hline\hline
			\end{tabular}
		\end{center}
	\end{table}
	
\begin{figure}[!htp]
\centering
\includegraphics[width=\linewidth, height=0.66\linewidth]{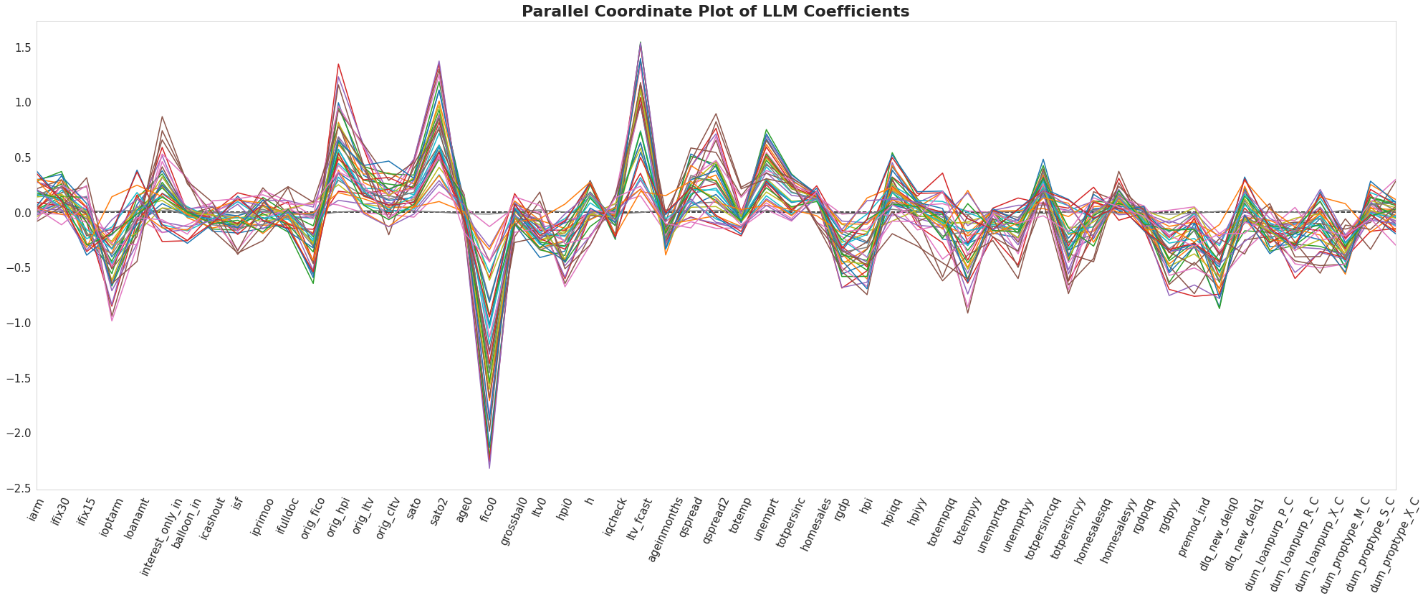}
\caption{PC plot for the Homelending data.}\label{homelending_PC}
\bigskip	
\centering
\includegraphics[width=\linewidth, height=0.85\linewidth]{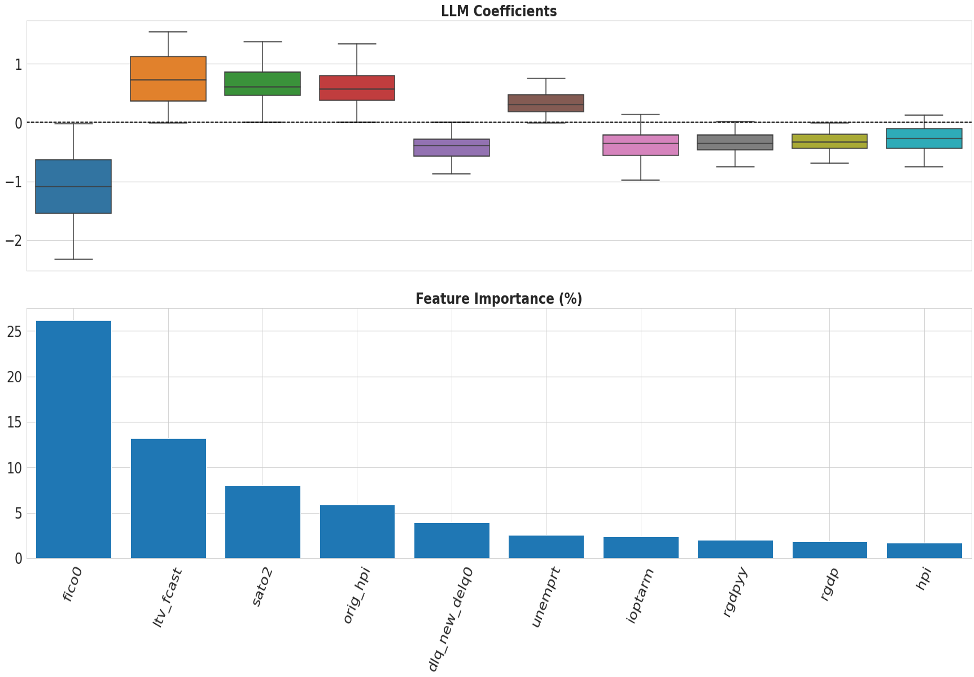}
\caption{Homelending case study: distribution of LLM coefficients (above); most important features (below).} \label{homelending_features_importance}
\end{figure}

\begin{figure}[!t]
		\begin{center}
				\includegraphics[width=0.9\linewidth, height=0.75\linewidth]{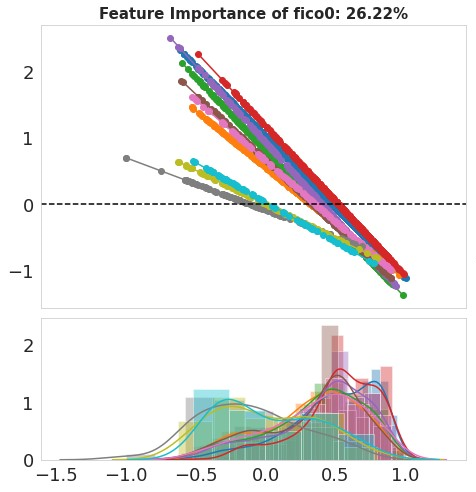}

				\includegraphics[width=0.9\linewidth, height=0.75\linewidth]{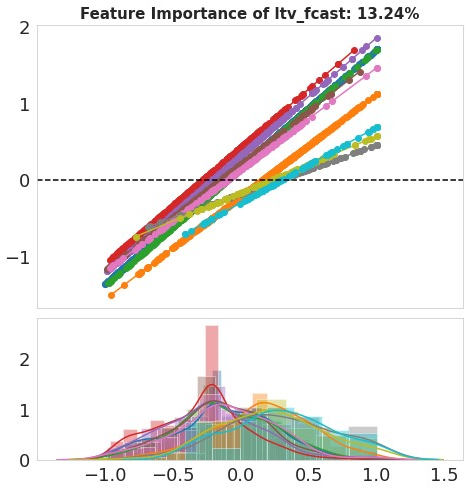}
			\caption{Profile of two most important variables. (a) FICO and (b) Forecasted Loan to Value (LTV).}
			\label{homelending_Profile}
		\end{center}
	\end{figure}

	Excluding the single instances and class, the coefficients of local linear 
	models are shown as a parallel coordinate plot in Figure~\ref{homelending_PC}. The top 10 most important variables and the distribution of their coefficients values are shown in Figure~\ref{homelending_features_importance}.  The local linear profile plots of the two most important variables, i.e., FICO and Forecasted Loan to Value (LTV), are shown in Figure~\ref{homelending_Profile}. The higher the FICO the less probability of default while the higher the LTV the higher probability of default, consistent with business intuition.
	
	Notice in Table~\ref{Homelending_llms} there are regions where the response means are either closer to 0 (non-default), 1 (default), or 0.5 (mix of default and non-default) with varying AUCs locally. To further enhance the model, we apply region merging particularly to address less reliable (small) regions where the sample size is small, particularly those regions with single sample size. Here we apply the merging algorithm proposed by \cite{sudjianto2020unwrapping}, which helps to reduce the ReLU-Net regions into three regions as shown in Table~\ref{homelending_merged}.
	
	The total AUC metrics of the merged regions (training AUC= 0.8532, testing AUC = 0.8388), in fact, are better than the original ReLU-Net (training AUC=0.8476, testing AUC=0.8316). This merging step shows that most of the regions created by ReLU-Net are redundant and can be replaced by three LLM that gives better predictive performance. Here, we applied penalized logistic regressions 
with regularization parameter $C=0.1$ for each merged region. 
	
\begin{figure}[!htp]
\centering
			\includegraphics[width=\linewidth, height=0.66\linewidth]{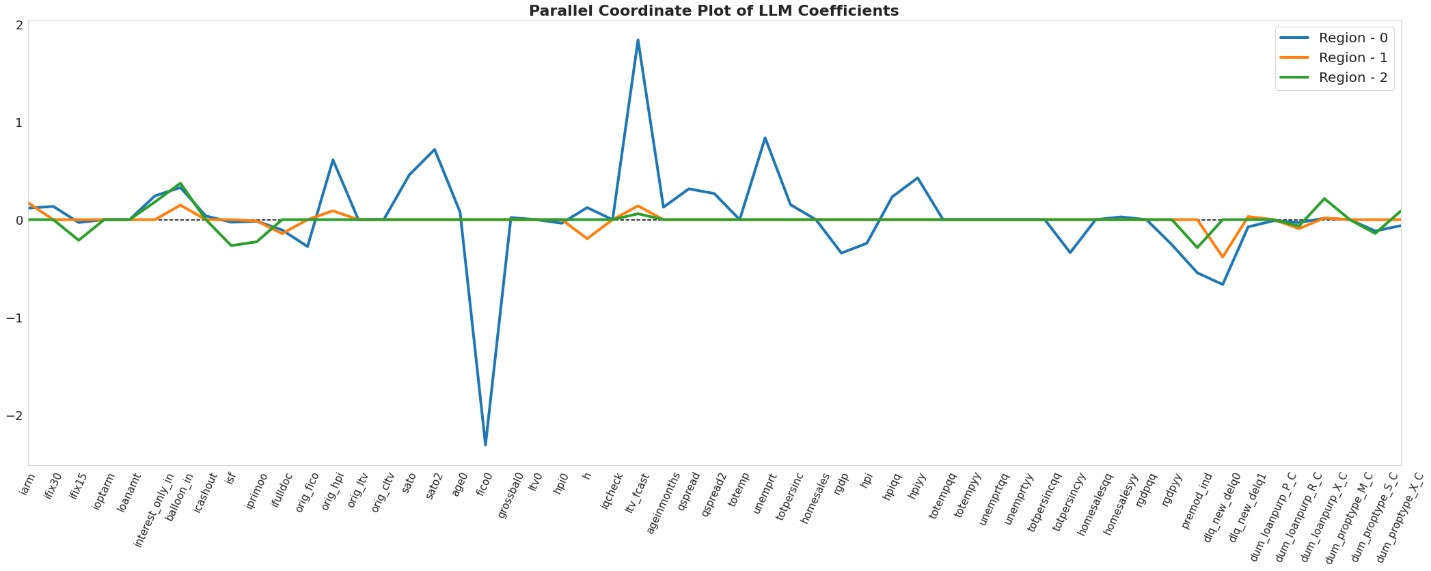}
			\caption{PC plots for the merged regions.}\label{homelending_merged_PC}

\bigskip\bigskip

\includegraphics[width=0.9\linewidth, height=0.7\linewidth]{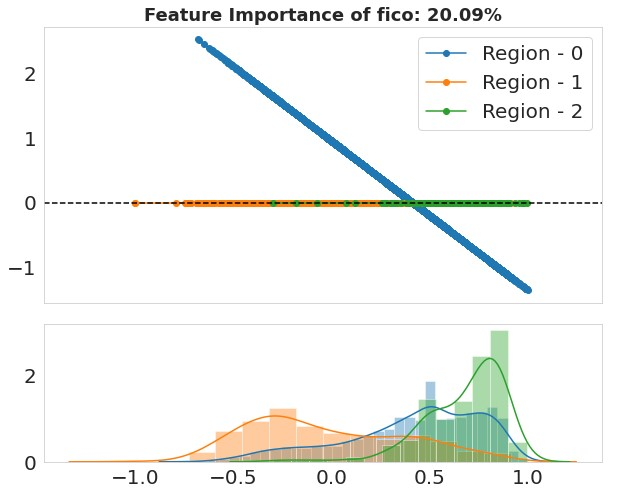}

\includegraphics[width=0.9\linewidth, height=0.7\linewidth]{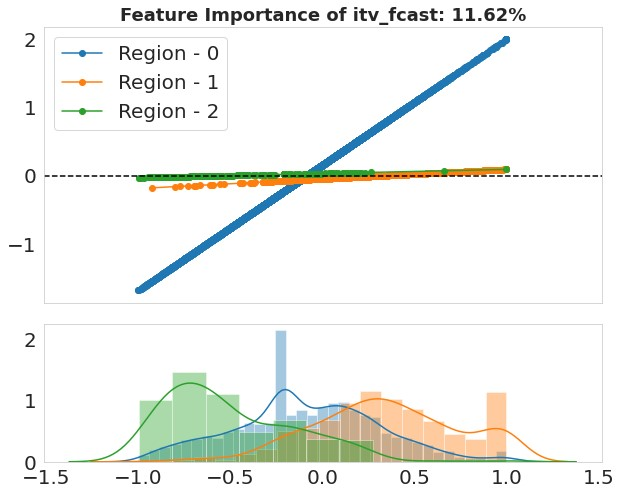}

\caption{Profile plots of the most important features.}  \label{homelending_merged_Profile}
\end{figure}

	Notice the results of the merging step created three distinct data segmentation (LLMs) with distinct characteristics. Region 0, the largest region with 85.2\% of the data contains mixed of default and non-default cases (response mean near 0.5). Region 1 consists of the majority default cases (90\%) where Region 2 consists of the majority non-default cases (85.4\%). The comparison between local and global AUCs also indicates the distinction among three regions as the local AUCs are better than the global AUCs. It is interesting to observe the parallel coordinate plot of the three LLMs and the corresponding profile of the most important variables shown in Figures~\ref{homelending_merged_PC} and~\ref{homelending_merged_Profile}, respectively. The profile plots indicate three distinct borrower characteristics: (1) High credit quality (high FICO, low LTV), (2) Middle credit quality (mid FICO and LTV), and (3) Low credit quality (low FICO, high LTV). The parallel coordinate plot provides contrast in terms of significant variables that drive the default event. FICO and LTV are important factors for the majority of the loans (Region 0). Delinquency status is the most important variable for the low credit quality (Region 1) where loans that are not currently delinquent will have a lower probability of default. Premod{\_}ind which is the indicator of loan origination before or after the financial crisis is important for the higher credit quality (Region 2). Loans originated post-financial crisis are better performing loans as they are subjected to a higher origination standard. It is also interesting to point out the effect of the ``h'' variable which is the time horizon of prediction, an equivalent concept to baseline hazard rate in the survival analysis model. For lower credit quality, the longer time horizon that the loans do not default the less likely the loans will default (i.e., decreasing hazard rate) while the high-quality loan is the opposite which is consistent with business intuition and credit theory from a stochastic process point of view.

	\begin{table}[!t]
	\setlength{\tabcolsep}{1pt}
\renewcommand{\arraystretch}{1.2}
		\begin{center}
			\caption{Summary of Three Merged Regions.} \label{homelending_merged}
			\begin{tabular}{cccccc}
				\hline\hline
				Region & Count & Response Mean & Response Std & Local AUC & Global AUC \\ \hline
				0      & 5602 & 0.460        & 0.498        & 0.831    & 0.854      \\ 
				1      & 747  & 0.904        & 0.293        & 0.680    & 0.584      \\ 
				2      & 219  & 0.146        & 0.353        & 0.987    & 0.736      \\ \hline\hline
			\end{tabular}
		\end{center}
	\end{table}

In the merging process above we retain the original networks for data partitioning and the final three LLMs are applied through a lookup table to map the regions from the original ReLU-Net to the merged regions. If a shallower network is preferred instead of using the original deep network, we may apply the flattening strategy proposed by \cite{sudjianto2020unwrapping}, which convert the final LLMs directly to an FL-Net (subject to retraining on the output weights). Here, we constructed a simple single hidden layer network with three nodes to represent the LLMs. The result comparison for the FL-Net is summarized in Table~\ref{homelending_Performance}. Note that the FL-Net has comparable performance as the Merged-Net and is better than the original ReLU-Net.
	
	\begin{table}[!t]
	\centering
		\renewcommand\tabcolsep{10pt}
		\renewcommand\arraystretch{1.4}
			\caption{AUC Performance Comparison of ReLU-Net, Merge-Net and FL-Net.} 	\label{homelending_Performance}
			\begin{tabular}{cccc}
				\hline\hline
				& ReLU-Net & Merge-Net & FL-Net \\ \hline
				Training & 0.8476   & 0.8532 & 0.8538    \\ 
				Testing  & 0.8316   & 0.8388 & 0.8368    \\ \hline\hline
			\end{tabular}
	\end{table}

\section{Conclusion}
 
In this paper we provide the design principles for developing high-performance IML models from feature selection and model architecture constraint perspectives. Such IML design principles are exemplified with our recent works on ExNN, GAMI-Net, SIMTree, and deep ReLU networks with local linear interpretability. Besides, we have also presented a real case study on home lending dataset, in order to demonstrate how to design a deep ReLU network with strong mechanism to evaluate model conceptual soundness equivalent to the statistical linear regression models. By using the Aletheia toolkit we have developed recently, it has been shown the deep ReLU networks can be easily interpreted with model diagnostics, and can be further simplified by the merging and flattening strategies. We hope that this work will provide a practical guide of developing inherently interpretable machine learning models in high risk applications in banking industry, as well as other sectors. 
	

\bibliographystyle{ACM-Reference-Format}
\bibliography{DesignIML}

\end{document}